# Using Learning-based Filters to Detect Rule-based Filtering Obsolescence

**Francis Wolinski [1,2], Frantz Vichot [1,3] & Mathieu Stricker [1,4]**

[1] Informatique CDC – Direction des Techniques Avancées
4 rue Berthollet, 94114 Arcueil, France
{forename.surname}@icdc.caissedesdepots.fr
[2] LIP6 Pole IA, 8 rue du Capitaine Scott, 75015 Paris
[3] LIMSI Langage et Cognition, BP 133, 91403 Orsay
[4] ESPCI Laboratoire d'Electronique, 10 rue Vauquelin, 75005 Paris

**Abstract**

For years, Caisse des Dépôts et Consignations has produced information filtering applications. To be operational, these applications require high filtering performances which are achieved by using rule-based filters. With this technique, an administrator has to tune a set of rules for each topic. However, filters become obsolescent over time. The decrease of their performances is due to diachronic polysemy of terms that involves a loss of precision and to diachronic polymorphism of concepts that involves a loss of recall.

To help the administrator to maintain his filters, we have developed a method which automatically detects filtering obsolescence. It consists in making a learning-based control filter using a set of documents which have already been categorised as relevant or not relevant by the rule-based filter. The idea is to supervise this filter by processing a differential comparison of its outcomes with those of the control one.

This method has many advantages. It is simple to implement since the training set used by the learning is supplied by the rule-based filter. Thus, both the making and the use of the control filter are fully automatic. With automatic detection of obsolescence, learning-based filtering finds a rich application which offers interesting prospects.

## 1. Introduction

Information filtering is taking on particular importance owing to the increasing volume of electronic documents available on networks. From the users' point of view, this technique is seen as a tool that reduces information overload. For the information suppliers, information filtering may also be used as a means of automatically creating a specialised wire from a more general one. To be attractive, a specialised wire has to bring a high added value to information that it broadcasts and to its presentation. Rule-based filtering is a technique that enables to limit precisely a flow of documents to those that deal with particular topics and to highlight parts of texts that have been used to select the documents. This technique requires an administrator to develop a set of rules for each topic.

For years, Caisse des Dépôts et Consignations (CDC) has produced such applications. To be operational, these applications require high filtering performances. Therefore, the administrator has meticulously to tune a set of rules for each topic. To do so, he builds up filtering rules that take into account features of topics and particularities of corpus. However, no matter how much care he takes to perform this task, one may observe that filters become obsolescent: their precision and their recall decrease over time. Therefore, to keep filters at a high level of performance, the administrator has to maintain them regularly in order to correct the errors that occur.

Number of filters put on line (100 topics) and rate of news (1,000 items a day) prevent the administrator from watching each of them individually. To reduce administrator's information overload, we have developed a method that automatically detects filtering obsolescence. It consists in making a learning-based control filter whose aim is to watch the initial filter. This method has many advantages. It is simple to implement since the training set used by the learning is directly supplied by the initial filter. Therefore the making and the use of the control filter are fully automatic.

Moreover, the administrator may work on parameters to adjust the rate of each control filter at a desired level. With this method, learning-based filtering finds a rich application which offers interesting prospects.

Section 2 presents applications that CDC has developed and the main features of rule-based filtering. Section 3 explains the rule-based filtering obsolescence phenomenon. Section 4 explains our method of making control filters using neural network techniques. Section 5 describes the architecture for controlling precision loss and recall loss. Section 6 gives qualitative results reached by this method. Section 7 places this method among the work that has been carried out in the domain of filtering, especially that relying on learning-based techniques and that dealing with adaptive filtering. Finally, Section 8 describes the future work that we are planning.

## 2. Industrial Use of Rule-based Filtering

This section presents the use of rule-based filtering that is done at Caisse des Dépôts et Consignations (CDC).

### 2.1 Rule-based Filtering at CDC

For years CDC has produced information filtering applications (Wolinski et al., 1998; Vichot et al., 1999). These applications process in real-time and enhance news wires supplied by Agence France-Presse (AFP). They are deployed for CDC's internal needs as well as for commercial purposes on the internet. To be operational, they rely on an effective rule-based filtering system which provides high performances (Landau et al., 1993; Vichot et al., 1997).

**2.1.1 Specialised News Wires** Several specialised information wires have been produced using this technique. A good example is *Mercure* which is available at `http://www.cdc-mercure.fr`. This French server counts more than 6,000 paying users among mayors, local authorities or Senators. This internet site includes 2 specialised filters for each French regions (general events and economic information) and about 30 headings organised according to local authority business (e.g., local democracy, regional development, decentralisation). The results of the filtering make it possible to consult news of each region or news about big companies operating in each region. For instance, it is important for a mayor to be acquainted as soon as possible with the fact that a local company is going to make mass redundancies or to relocate. In this application, titles of news items are listed with the name of the subdivision or of the firm that is involved (Figure 1). This information allows users to read only local news items that interest them.

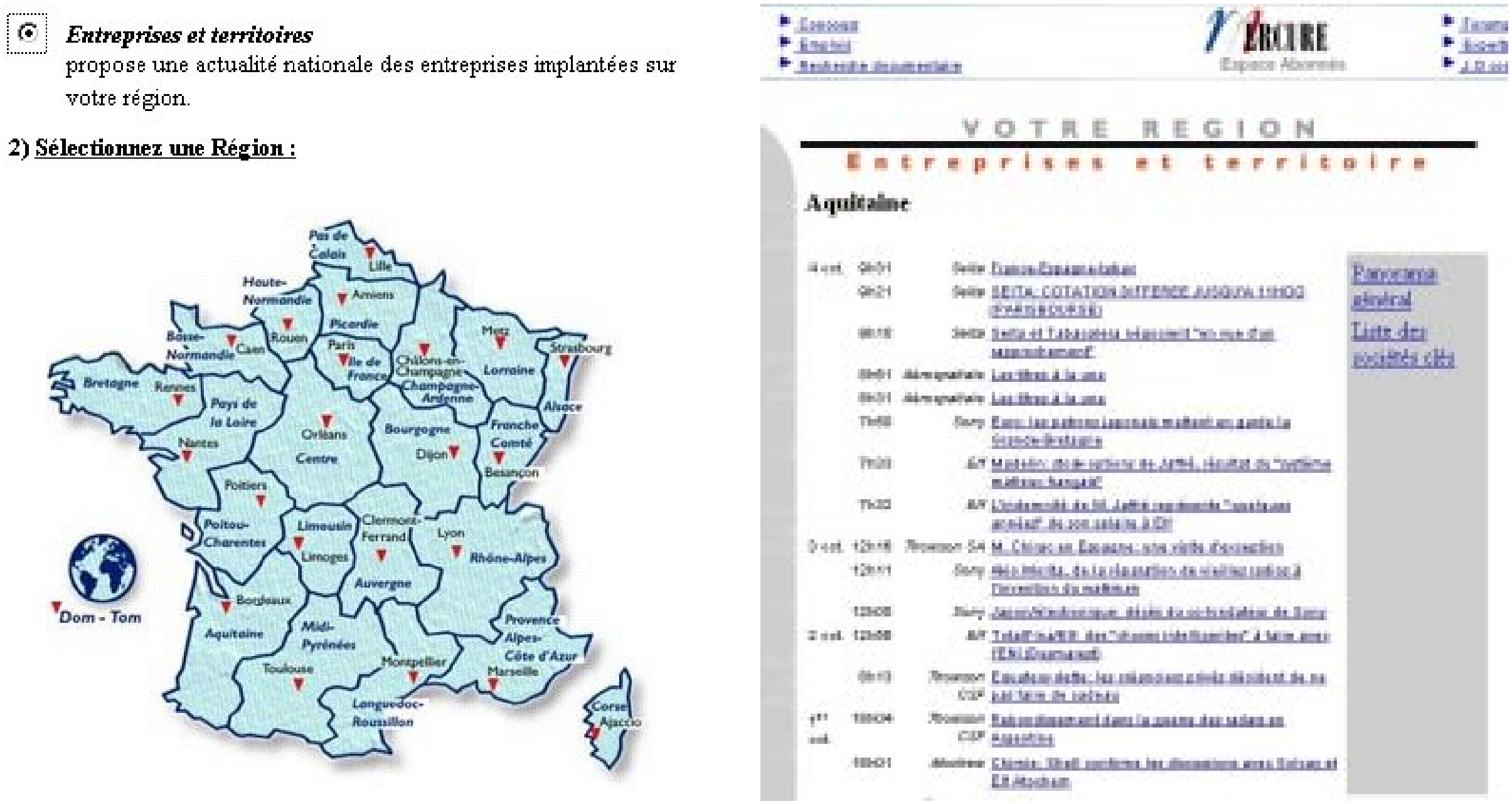

Figure 1: Specialised news wires

**2.1.2 Specialised Data Wires** CDC has also produced specialised data flows for managers of stocks portfolios (Vichot et al., 1999). Each data flow concerns only shareholdings on companies of a given portfolio. A user interface enables the manager to view simultaneously a news item about shareholdings and the corresponding part of his portfolio database (Figure 2). The relevant parts of the text are highlighted. The graph of shareholdings is limited to the companies that are effectively mentioned in the document. In this way, the manager is able to compare information provided by the news and the information already in his database.

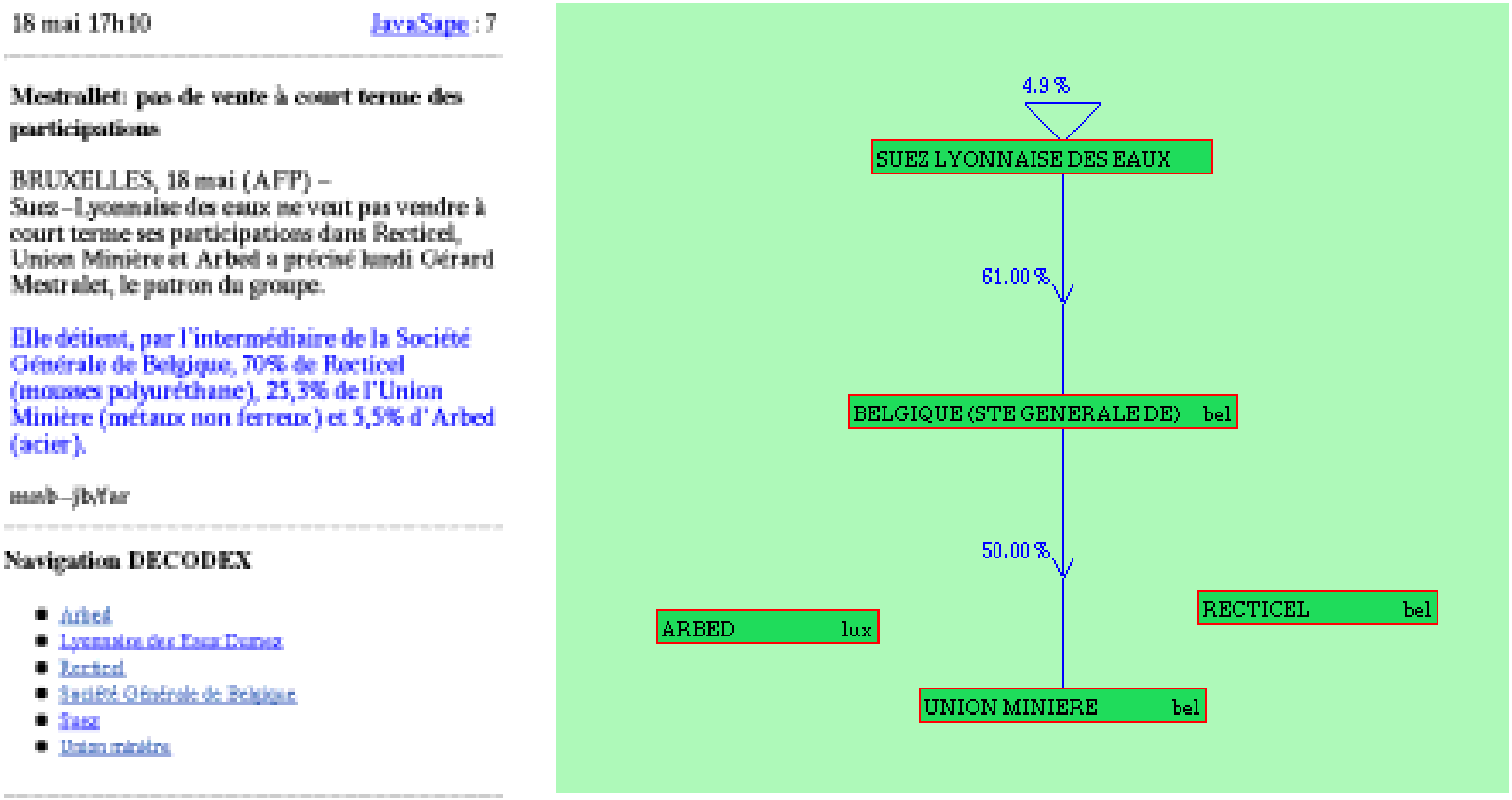

Figure 2: Specialised Data Wire

## 2.2 Rule-based Filtering Overview

All these applications rely on a rule-based filtering technique. This technique consists in locating specific terms that are used to mention the concept that is being sought (Figure 3). A filter uses a set of complex rules, developed by an administrator, which make it possible to limit the semantic scope of terms. To develop a rule base an administrator has to find a trade-off between users interests and realities of corpus. There is a way to define rules of extreme precision that rely on particular syntactic relations or internal structure of documents. For instance, we have described (Landau et al., 1993) how syntactic relations can be used to detect complex concepts (e.g., companies shareholdings, VIP appointments, announcements of company results) and (Wolinski et al., 1995) how local and global contexts of terms can be used to disambiguate proper names (e.g., homonyms, acronyms).

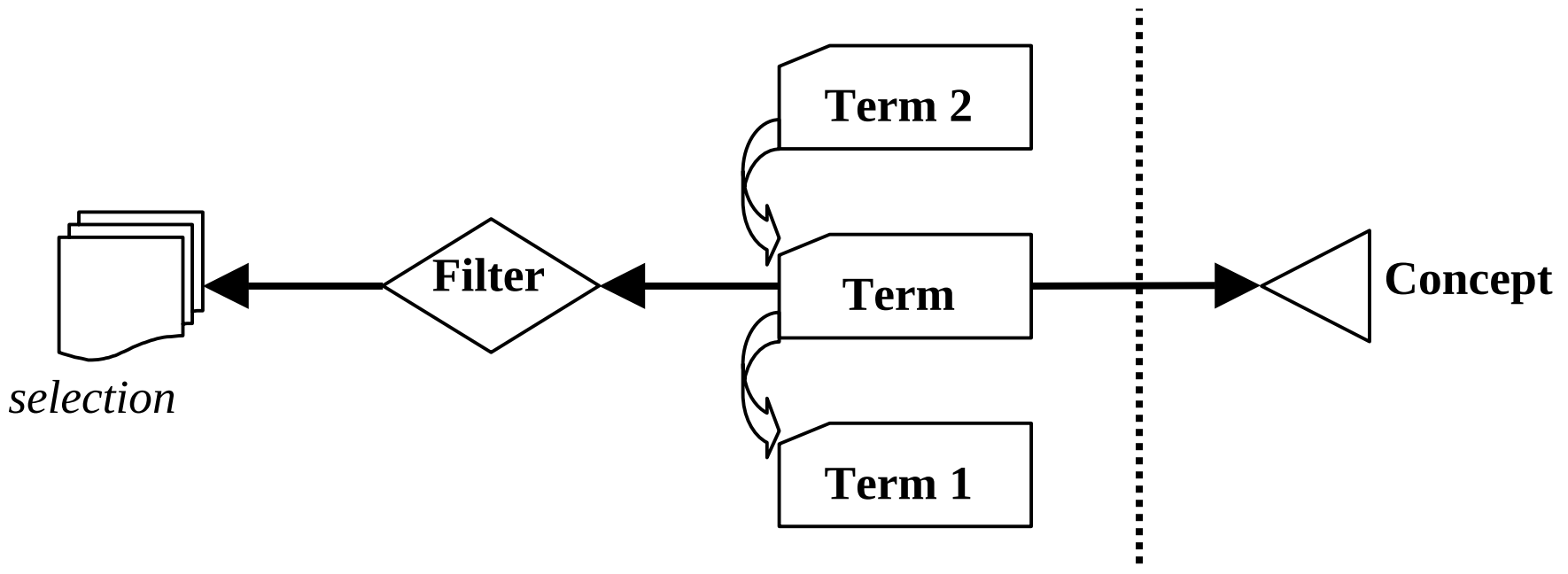

Figure 3: Rule-based filtering

This filtering technique uses a limited number of rules. These rules refer to a pattern involving a restricted set of words that have been chosen by the administrator. An inference engine triggers the rules in forward chaining when the words mentioned in a rule are simultaneously present in a single sentence or in the whole text and are linked with the syntactic relations expressed in the pattern. However, when filtering a corpus such as news releases, that deal with passing, varying and fluctuating events, the performances depend on the subjects that are dealt with and on the expressions that are used by journalists. In this case, rule-based filtering may be effective but it remains fragile, especially over time. The main problem of this technique is the obsolescence of filters.

## 3. The Obsolescence of Rule-based Filtering

Over time, the performance of a rule-based filter decreases. This slow degradation is more noticed rather than precisely measured. Users of filters sometimes point out that a document has been wrongly selected by a filter or conversely that they have not been informed of a fact which they discovered from another source. These accidents are a real nuisance for users. When a document is wrongly selected, users may think that the filtering system is not efficient. When information is missed, users may think that the filtering system (or the information source) is not reliable.

The obsolescence of a filter *over time* has two causes. Firstly, some terms that are used become polysemic. Secondly, some concepts that are mentioned become polymorphic. In this paper, these two phenomena are called *diachronic polysemy* and *diachronic polymorphism*.

### 3.1 Diachronic Polysemy

Polysemy is the ability of a given term to refer to different concepts. Over time, any term is potentially polysemous. When polysemy does not appear, there is no problem. But, when a current event deals with a concept referred to by the very same term as the one selected by a filter, diachronic polysemy appears and filter precision decreases (Figure 4).

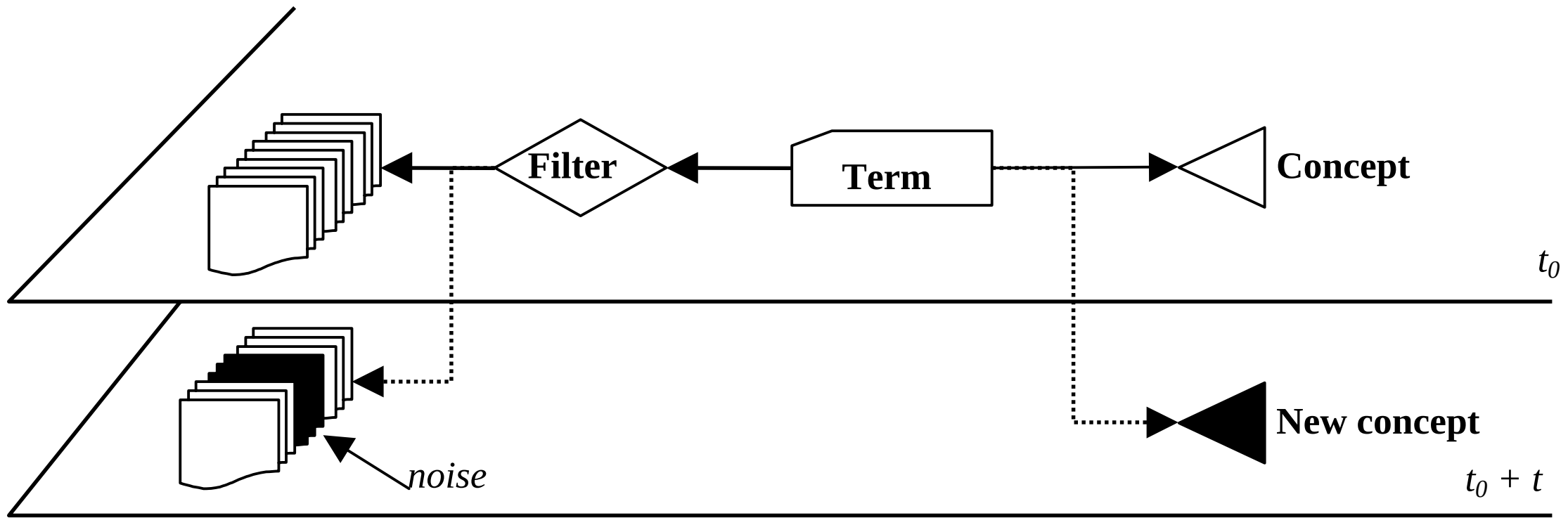

Figure 4: Diachronic polysemy and precision loss

Proper nouns make up a large field of polysemic terms. This phenomenon is due to frequent homonyms. For instance, *Saint-Louis* refers simultaneously to a French company, an American city, a Parisian hospital, or again a king of France. Thus, a filter dedicated to the company *Saint-Louis* has to disambiguate the different meanings of this term. In fact, the set of the different meanings of an ambiguous term is potentially infinite. For *Saint-Louis* we have also a small city in east of France, another French company specialised in crystal glass manufacture, a street name referring to the king, etc. Another cause of polysemy is given by acronyms which are naturally shared. For instance, *CDC* fits with several organisations : *Caisse des Dépôts et Consignations*, *Center of Disease Control*, *China Development Corp.*, etc. In this case, diachronic polysemy is due to existing concepts which were not in the scope of the newswire and which suddenly appear or to recent concepts (e.g., new companies, organisations or products) which suddenly arise.

When a term acquires a new meaning, a filter selects the documents that contains this meaning even though they are not related to the topic. Therefore polysemy involves a loss of precision. To limit polysemy effects, a rule-based filter contains disambiguation techniques. They enable in some cases to distinguish precisely the concept that is referred to or at least to reduce the set of concepts that are potentially referred to (Wolinski et al., 1995). For instance, a disambiguation rule may be: if the word *maire* (*mayor*) has a proper name in a possessive phrase then the proper name is a location. So, the sentence "le maire de Saint-Louis" (the mayor of Saint-Louis) does not deal with the company. However, these techniques require that these cases be previously listed which cannot be done when a new meaning appears for the first time.

### 3.2 Diachronic polymorphism

Polymorphism is the ability of a given concept to be referred to by different terms. Over time, terms that refer to concepts are subject to changes that result from language evolution. When a new term is used to refer to a given concept, diachronic polymorphism appears and filter recall decreases (Figure 5).

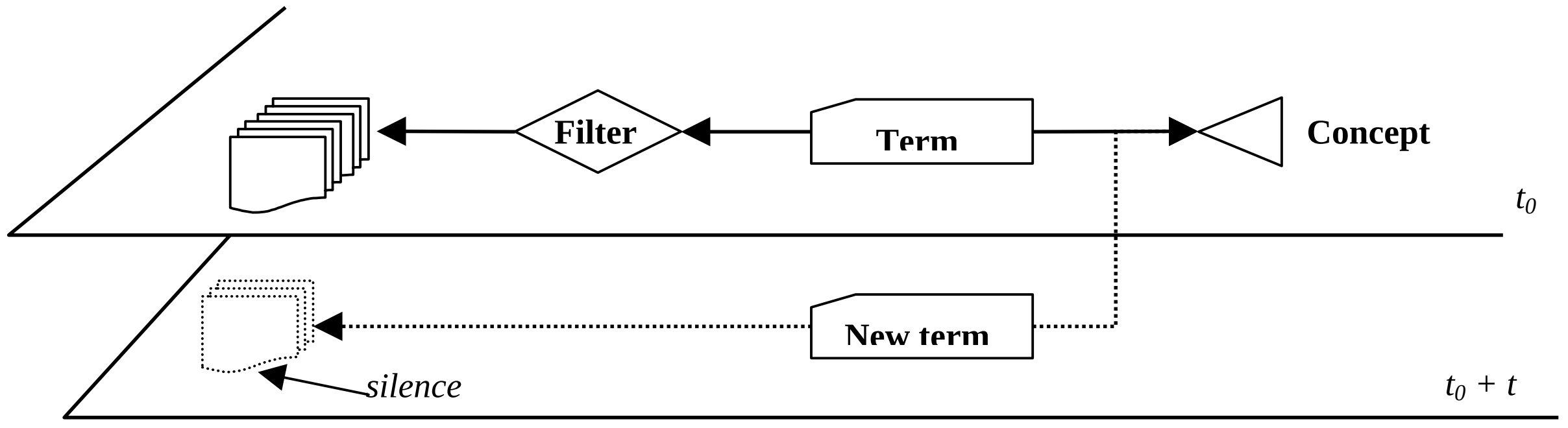

Figure 5: Diachronic polymorphism and recall loss

Proper nouns make up a large field of polymorphic concepts as well. This phenomenon is due to frequent variations of forms that occur through human writing. Examples are various since one may find misspellings (*Eltsine*, *Elstine*), abbreviations (*Société Générale*, *SocGen*), translations (*Pékin*, *Beijing*), name changes (*Compagnie Générale des Eaux*, *Vivendi*) or again metaphors (*IBM*, *le groupe d'Armonk*). Diachronic polymorphism is due to existing terms which were not in the scope of the newswire and which suddenly appear or to variations of terms which suddenly arise.

When a concept takes on a new form, a filter rejects the documents that contain it even though they are on the topic. Therefore polymorphism involves a loss of recall. To limit polymorphism effects, pattern matching techniques may be used in some cases when both terms are used in the same document (Wolinski et al., 1995). But these techniques are inefficient when the new form appears alone for the first time.

### 3.3 Some More Complex Examples

The concepts that are filtered are subject to state changes whose effect is also diachronic polymorphism (Figure 6). Over time, a concept may change into another, or even disappear in favour of a larger concept or smaller ones. These changes refer to the definition of topics, that is to say, to the fit between users interests and information flows that are supplied to them. Daily life provides numerous examples of this phenomenon: states divided or reunified, companies absorbed or merged, variable sets of companies in financial index, etc.

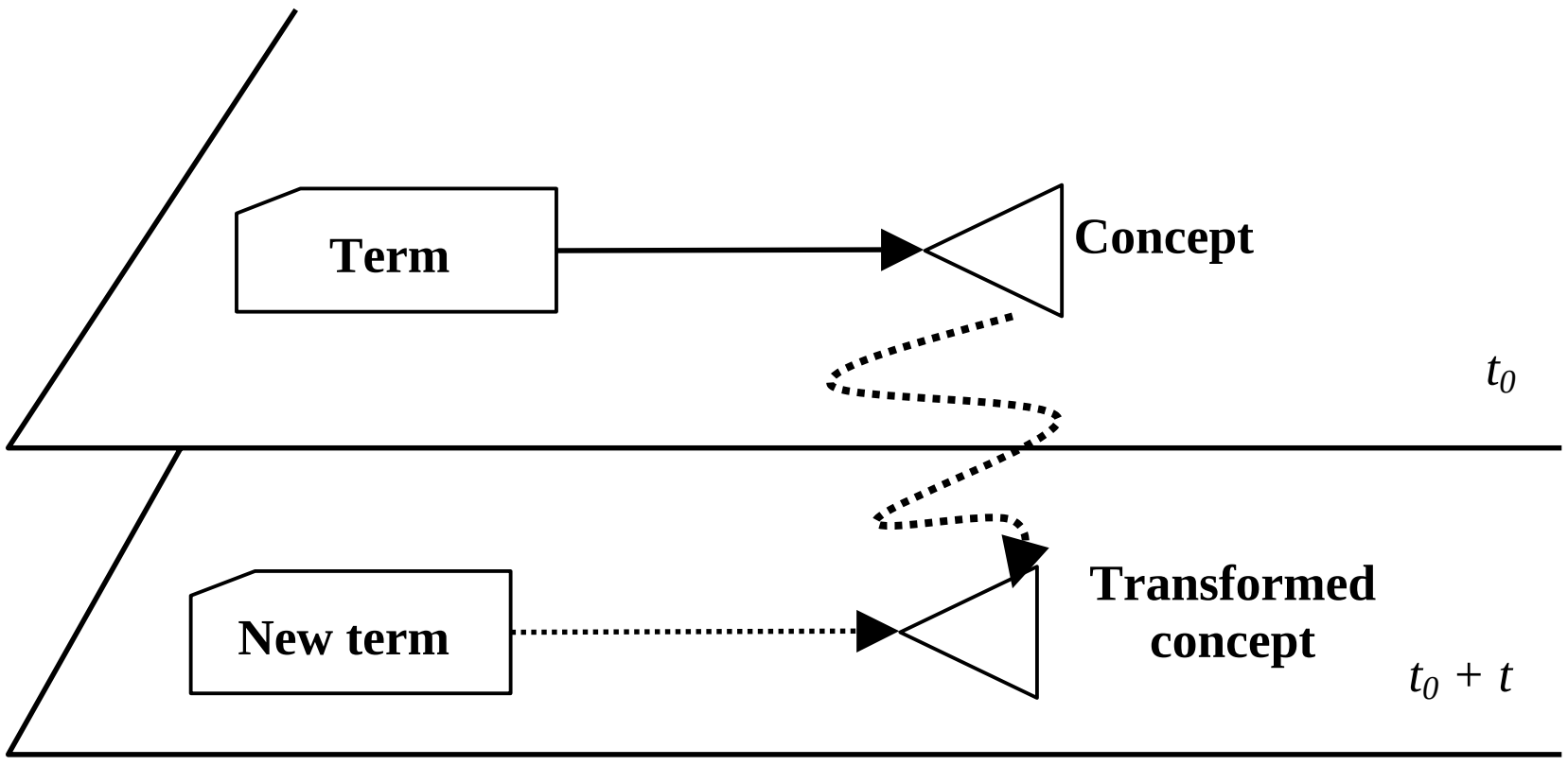

Figure 6: Diachronic polymorphism: concept and term drifts

Polysemy and polymorphism may even interact. For instance, a French department store, *Bazar de l'Hôtel de Ville*, is also referred to by the well-known acronym *BHV*. Recently, two German banks, *Bayern HypothekenBank* and *Vereinsbank*, have merged. The product of this merger is now called *BayernHypo-Vereinsbank*. Newspaper journalists shorten this name by using *BHV* as well. In their mind, no confusion is possible: a news item deals with the one or the other company. But at the rule-based system level, the acronym is now shared by two different companies and it has to be disambiguated.

## 4. Introducing Learning-based Filtering

This section presents our method to build a learning-based filter from the results of a rule-based one.

### 4.1 Learning-based Filtering

Unlike rule-based filtering, which relies on the detection of precise terms referring to each concept, learning-based filtering generalises configurations of terms that are shared by documents which deal with the same topic (Figure 7). Learning-based filtering may use any term which has been selected by an automatic processing of a large number of examples. In a sense, such a filter learns the terminological context of documents which share the same topic.

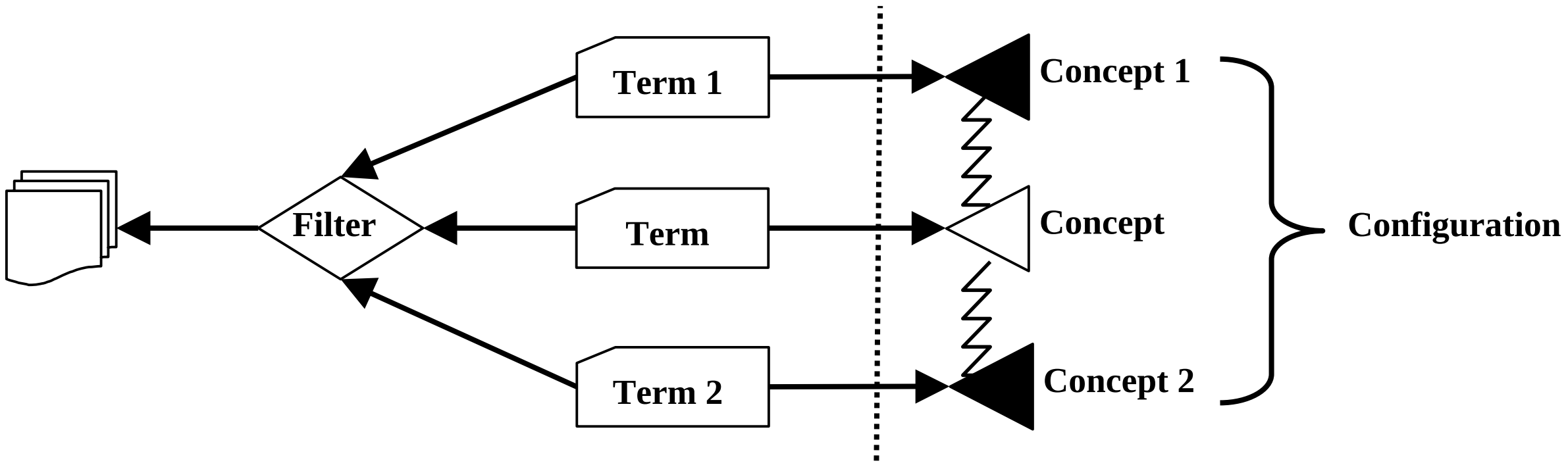

Figure 7: Learning-based filtering

In a way, learning-based filters are context sensitive whereas rule-based filters are more pattern oriented. Therefore, learning-based filters are more robust to changes in topics or in terms since the selection of a document always relies on the co-occurrence of several terms of the domain.

### 4.2 Training a Learning-based Filter with Rule-based Filtering Results

Let *F* be a rule-based filter, called *initial filter*, our method automatically builds up a learning-based filter *F'*, called *control filter*. After that, supervision of the initial filter consists in making a differential comparison between its own outcomes and those of the control filter (see Section 5).

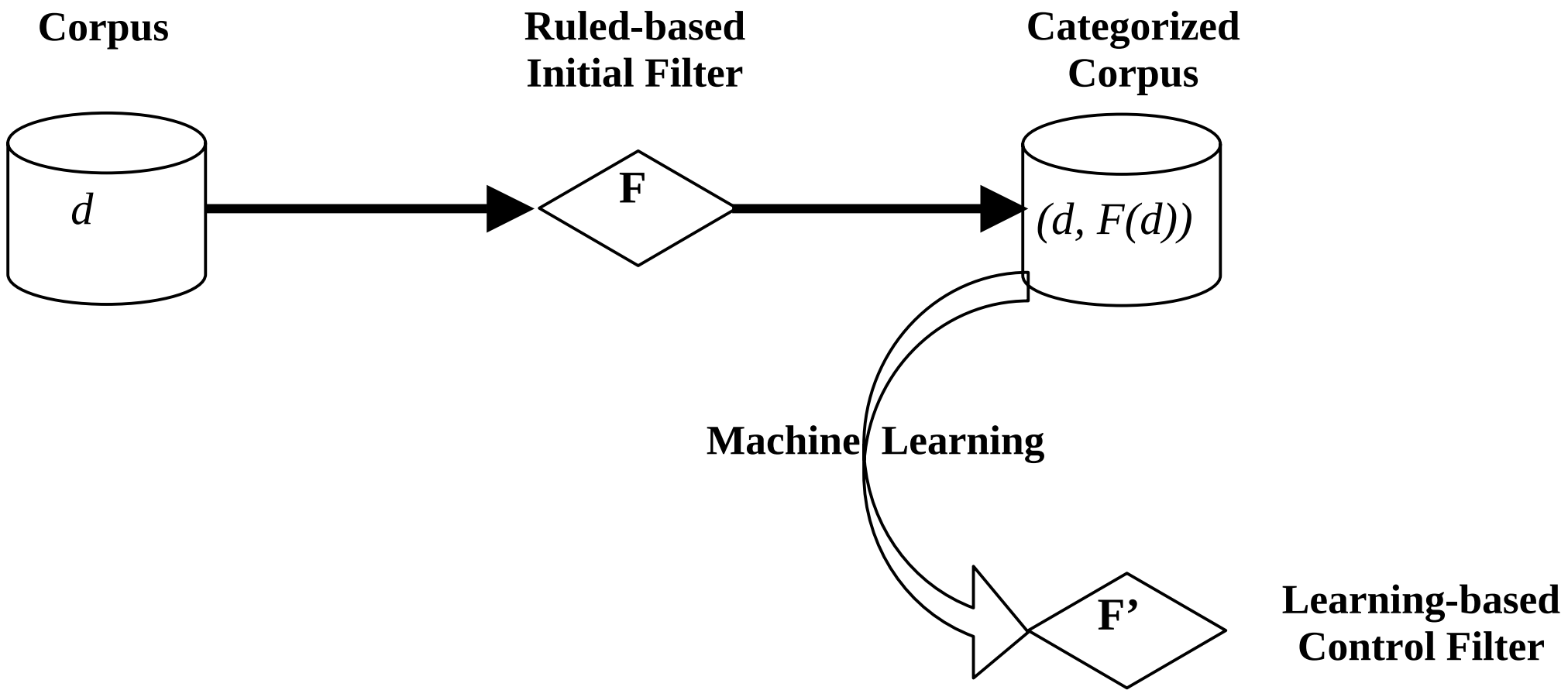

Figure 8: Training a filter with the results of another

The control filter is made by using a learning technique (Figure 8). One of the problems with learning-based techniques is to gather a significant base of examples manually. In our case, the learning base is automatically given by the documents that have been categorised by the initial filter, i.e. the documents that have been selected to be on the topic processed by the filter. If the initial filter is rather verbose, the learning base is built by selecting at random a subset of the documents selected in a given period. If the initial filter is rather mute, the learning base is built by collecting all documents selected in a large historical period. Hence, the problem of making the learning base simply vanishes.

The control filter is built by using the results of the classification performed by the initial filter. Thus, it is similar to the initial one. That is to say, a document correctly selected by the initial filter should be rated close to *1* by the learning-based filter, and conversely, a document correctly rejected by the initial filter should be rated close to *0* by the learning-based filter. We will see in Section 5 how this similarity between the two filters is used to detect obsolescence of the initial one.

### 4.3 Implementation of a Control Filter with a Neural Network Classifier

Our method assumes that the performances of the initial filter are high enough during the learning of the control filter that the latter is itself of good quality. The making of the control filter does not presume anything about the learning method that is used. Any method able to learn text categorisation from positive examples (given by the documents selected by the rule-based filter) and even negative examples (given by the documents rejected by the rule-based filter) could be a good candidate to detect rule-based filtering obsolescence.

We have tested the learning technique that we have used elsewhere for TREC-8 routing. This technique relies on a neural network classifier. It consists in the training of a neural network for each of the filters to be controlled. In the following subsections, we give an overview of this learning-based filtering technique. The reader can refer to the TREC-8 proceedings for a complete description and evaluation of the method (Stricker et al., 1999).

**4.3.1 Term Selection** The goal of the term selection is to define, for each topic, a vector of terms that will represent each document. The choice of these terms must be done very carefully since the quality of the filter relies heavily on this choice. Texts of the learning base are tokenised into single words. Terms occurring with a high frequency or a low frequency are simply discarded whereas remaining terms are ranked using the Gram-Schmidt orthogonalisation technique (Chen et al., 1989). Finally, a statistical criterion makes it possible to retain only discriminant terms, i.e. terms which are relevant according to the topic. The result is a vector of terms which are specific to each topic. For instance, for the 50 topics of TREC-8 routing, the average length of the vector were 25 terms.

**4.3.2 Neural Network Training** For each topic, the term frequencies of the vectors defined above are used as inputs of a neural network. Firstly, we had chosen the simplest architecture for the neural network i.e. a simple unit with a hyperbolic tangent function. We are now testing more complex neural networks able to handle structural relationships between words. The classifier is trained by minimising the mean square difference between the desired output and the actual value of the classifier on the training set (supplied by the results of the initial filter as explained in Section 4.2). The desired value is *1* if the document is relevant and *0* if the document is not relevant.

**4.3.3 Neural Network Classification** After the training step, the neural network is able to process each document and to compute the probability of relevance. The following section explains how the neural network classifier is used to control the rule-based filter.

## 5. Using a Learning-based Filter to Detect Obsolescence

This section explains how the learning-based filter is used to control a rule-based one. The architecture for detecting precision loss and that for detecting recall loss is symmetrical.

As we have seen, the obsolescence of a filter has two causes: diachronic polysemy and diachronic polymorphism. We explain in sections 5.2 and 5.3 how we deal with both of them. First, we shall introduce some mathematical notations that are used in these sections in order to facilitate our presentation.

### 5.1 Notations

Let *F* be a rule-based filter, it is a function valued in the set *{0,1}*.
Let *d* be a document, *F(d) =1* if *d* is on the filtered topic and *F(d)=0* if not.
Such a filter may be seen as a binary classifier which associates to a document *d* the category *F(d)*. We denote by *1-F* the inverse filter of *F*.

Let *F'* be a learning-based filter, it is a function valued in the interval *[0,1]*.
Let *d* be a document, *F'(d)* is the probability that the document *d* is on the filtered topic.
In fact, such a filter must be associated with a threshold (taken between *0* and *1*) to take decisions about selecting a document or not.

### 5.2 Detecting Precision Loss

We have seen that diachronic polysemy of terms involves precision loss. Therefore the problem is to detect this polysemy. When a term refers in a unique way to a single concept, we may define the terminological context of documents that use it. Thus, if this term takes on a new meaning, it will be used in a truly different context (Figure 9).

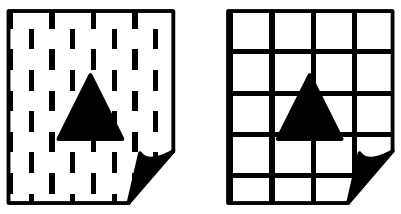

Figure 9: Polysemy: same term, different contexts

To detect this case, we evaluate by the control filter *F'* the documents that have been selected by *F*. By applying a threshold $s^-$ to the results of *F'*, we extract the documents that should not be selected by *F*. If for a document *d* we have *F(d)=1* and $F'(d)<s^-$ then polysemy may be suspected. In fact, supervising the precision of a filter *F* is done by applying the filter *F'* on the outcomes of *F* and by composing it with a threshold function (Figure 10).

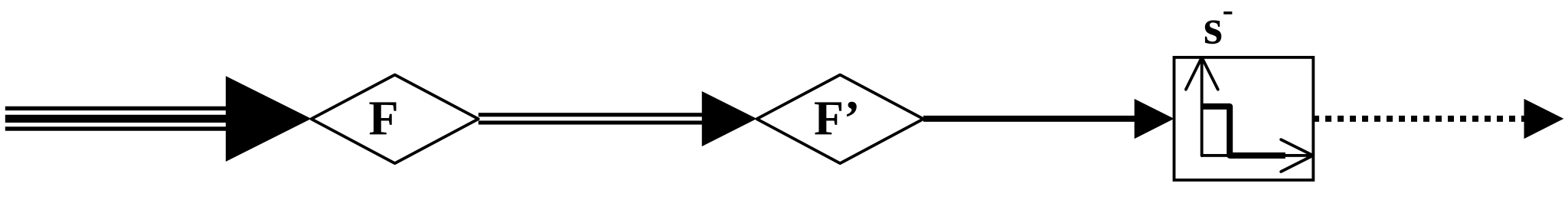

Figure 10: Architecture to detect precision loss

The threshold is set empirically by the administrator. The higher it is, the more verbose the control filter will be. Intuitively, the documents *d* where *F(d)=1* and where *F'(d)* is close to *0* are most likely assumed to be some noise of filter *F*. The idea here is that the administrator is given the opportunity to tune individually each control filter according to the care he devotes to checking each initial filter.

### 5.3 Detecting Recall Loss

We have seen that diachronic polymorphism of concepts involves recall loss. Therefore the problem is to detect this polymorphism. When a concept is referred to in a unique way by a single term, we may again define the terminological context of documents that use it. Thus, if this concept takes on a new form, it will be used in a very similar context (Figure 11).

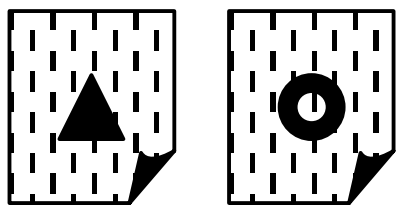

Figure 11: Polymorphism: different terms, same context

To detect this case, we evaluate by the control filter *F'* the documents that have been rejected by *F*. By applying a threshold $s^+$ to the results of *F'*, we extract the documents that should be selected by *F*. If for a document *d* we have *1-F(d)=1* and $F'(d)>s^+$ then polymorphism may be suspected. In fact, supervising the recall of a filter *F* is done by applying the filter *F'* on the outcomes of *1-F* and by composing it with a threshold function (Figure 12).

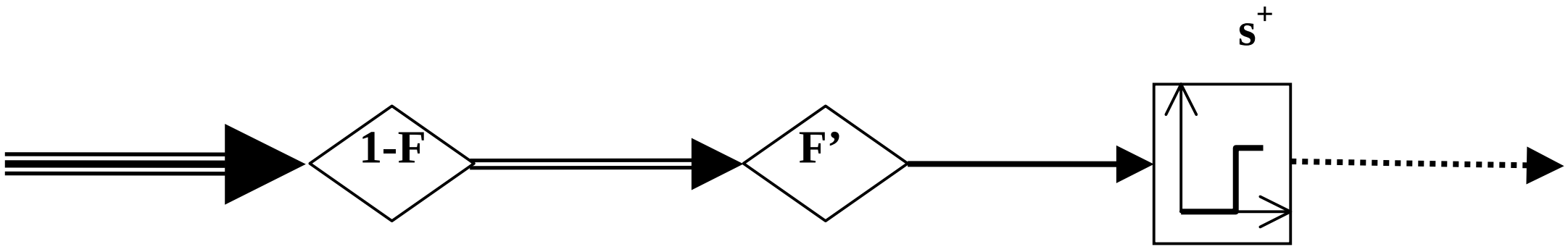

Figure 12: Architecture to detect recall loss

Here again, the threshold is set empirically by the administrator. The lower it is, the more verbose the control filter will be. Intuitively, the documents *d* where *1-F(d)=1* and where *F'(d)* is close to *1* are most likely assumed to be some silence of filter *F*.

## 6. Results and Discussion

We have tested this method on two kinds of filters. Those dealing with a specific proper noun (e.g., Caisse des dépôts et consignations) and those dealing with a larger domain (e.g., information highways). The following table summaries the qualitative results that were obtained with these two kinds of filters (Table 1).

| | **Diachronic Polysemy** | **Diachronic Polymorphism** |
|---|---|---|
| **Proper Name Filter** | Homonym<br>New Context | Synonym<br>Similar Context |
| **Domain Filter** | Simple Reference<br>Domain Extension | Emerging Vocabulary<br>Related Domain |

Table 1: Synthesis of qualitative results

### 6.1 Detection of precision loss

Our method effectively detects precision losses caused by homonyms. For instance, the rule-based filter will not reject news items about *Caisse de dépôts du Maroc* or *Caisse de dépôts du Québec*, organisations independent of CDC, until specific rules for dealing with these cases are implemented. The rule-based filter rejects such news items because the context is generally quite different from CDC's. The method also detects simple references; that is to say, news items in which the concept that is sought is simply quoted but is not the main subject. For instance, a news item like "*Le rapport Starr sur Internet*" (The *Starr Report on the internet*) will be selected by the rule-based filter but rejected by the control one. In fact, a news item which simply mentions internet is not necessarily of great interest in the field of information highways.

However, the method may produce new contexts; that is to say news items effectively dealing with a company but within contexts radically different from the standard ones. For instance, a news item on a recent law simply mentioning CDC will be selected by the initial filter but rejected by the control one. Finally, the method also detects domain extensions; that is to say news items dealing with a large subtopic of the filtered topic. For instance, as security on the internet has become a real subject on its own, news items dealing with security which are selected by the initial filter may be rejected by the control one. This is due to the extremely rich vocabulary which comes with this specific subtopic (e.g., encryption, firewall, hackers, cybersquatting) and which produces a radically different context.

### 6.2 Detection of recall loss

Our method effectively detects recall losses caused by synonyms. For instance, sometimes the location of CDC in Paris (*Rue de Lille*) is used instead of CDC, especially when a VIP of CDC is mentioned. The rule-based filter will reject news items dealing with *Rue de Lille* until a specific rule is implemented. The rule-based filter selects such news items because the context is usual to CDC. The method also detects emerging vocabulary. The internet gives a lot of significant examples of such a phenomenon: new concepts, new products, new web sites and new companies come out everyday.

However, the method may produce similar context; that is to say news items dealing with companies that share the same context of the one that is sought. For instance, the business of *Caisse d'Epargne* is rather close to that of CDC, moreover both companies have along partnership together. Some news items dealing only with *Caisse d'Epargne*, thus rejected by the initial filter, may be selected by the control one. Finally, the method also detects related domains; that is to say news items dealing with a topic rather close to the filtered one. For instance, news items on mobile phones and m-business or TV and multimedia which are rejected by the initial filter may be selected by the control one.

### 6.3 Discussion

The qualitative results expressed above show at the same time the efficiency and the limits of the method. We are aware that our method whose aim is to detect noise and silence in the initial filter produces noise and silence in the control filter as well. However, through their construction, control filters give the administrator of rule-based filters access to borderline documents which were unreachable until now. Using the thresholds of the control filters, the administrator may choose their trigger level according to the rate that he needs. Our opinion is that with control filters the administrator becomes more efficient. He is more promptly alerted (and not by users) when errors occur. And moreover he gets useful examples of the mistakes made by rule-based filters which are directly exploitable to correct them.

We are aware of the lack of quantitative evaluation of our method. The fact is that performances of a control filter should not be computed with the same tool than the one used for an initial filter. In our method, a rule-based filter is intended for a final user. The user is interested in receiving all information he need and only information he needs. The performance of such a filter may be computed in terms of precision $p$ and recall $r$ by using F-measure: $f = 2\,p\,r / (p + r)$ defined in (Lewis & Gale, 1994). But, a control filter is intended for an administrator. The administrator is not really interested in receiving all potential diachronic polysemy or polymorphism because he will not have time to correct everything anyway. The administrator is interested in receiving significant or recurrent mistakes. Therefore, the rating of each news item provided by the control filter is used to sort the news items which are presented to the administrator. Thus, the system enables the administrator to deal first with the most significant cases (according to the control filter). Otherwise, by coupling the results with a clustering system, the most frequent mistakes can be displayed as well.

## 7. Related Work

Among information filtering and text categorisation techniques, learning-based methods have appeared in recent years. First, they rely on the selection characteristic features of documents. Second, they use different learning techniques such as regression models, nearest neighbour classification, neural networks, Bayesian probabilistic approaches, decision trees, etc. The reader can find in review (Yang & Lui, 1999) these different methods used for information filtering or text categorisation with their respective performances.

The main problem with these learning methods is that learning requires a significant base of labelled examples. Some works try to minimise the efforts needed to build up such a base (McCallum & Nigam, 1998). One (Stricker et al., 2000) describes a technique that relies on a search engine linked to a neural network in order to enable the administrator to build up the learning base with a minimum of work. To make a control filter from an existing filter, we have seen that the building of a learning base is no longer a problem. The learning base is automatically given by the documents that have been categorised by the initial filter.

To follow the inevitable changes that occur in an information wire, it is possible to resort to adaptive filtering techniques. The idea is to detect a change in the performance of a filter in order to update it automatically with a learning technique. Our approach shares with adaptive filtering the problem of automatically detecting changes in an information flow, but in our case, the update of filters is always done manually by the administrator. To detect such changes, some authors resort to total or partial user feedback (Klinkenberg & Renz, 1998). Our method prevents the need for the full outcome of each filter to be watched by the administrator.

Obsolescence of filters has also been tackled from the point of view of dynamic evolution of document topics (Lanquillon, 1999). This work has proved that detecting a change in a corpus was possible without user feedback, but with a temporal comparison of the outcomes of the very same filter. Our method processes a differential comparison of the outcomes of an existing rule-based filter with those of a learning-based control filter. It detects evolutions in polysemy of terms that are used and in polymorphism of concepts that are mentioned.

Finally, for the filtering task TREC has defined three subtasks: batch, routing and adaptive filtering (Hull, 1998). In batch filtering, the system decides either to accept or reject each document, while in routing, the system computes a probability for each document. Our method consists in supervising a batch filtering system with a routing one. Our method should also be compared to news filtering learning systems that observes users “over the shoulder” such as NewT (Maes, 1994). In this case, the differences between the selection of the system and the selection of the user are used to improve the filtering system automatically. In our case, the differences between the selection of both initial and control filters are used by the administrator to improve the system manually. In our case no feedback from users is needed.

## 8. Conclusion and Future Work

In this paper, we have presented a method that automatically detects rule-based filtering obsolescence. It reveals precision losses caused by diachronic polysemy of terms and recall losses caused by diachronic polymorphism of concepts. This method relies on the creation of a control filter that doubles the initial filter. This control filter is made by using learning techniques on the outcomes of the initial filter. Supervising a filter consists in processing a differential comparison of its outcomes with those of the control filter. News items selected by the initial filter but rejected by the control one are supposed to decrease precision. Conversely, news items rejected by the initial filter but selected by the control one are supposed to decrease recall.

Another application of our method is to bring assistance to the administrator for developing new rule-based filters. The idea is to develop a first version of a rule-based filter, then to build a control filter with a learning base, and then to evaluate the first version with the control filter on a test base and finally to improve the rule-based filter using the mistakes that have been found. This iterative development technique of rule-based filters opens up the way to new uses of learning techniques in information filtering.

Finally, our method for tackling with the problem of detecting precision and recall losses has showed that rule-based and learning-based filtering techniques are quite complementary. One idea could be to try to combine those techniques in order to produce a better filtering system. In our architecture, the control filter is a kind of a *meta-filter* which should be given the opportunity to correct the initial filter continuously. Such a reflexive architecture could be connected to works in Artificial Intelligence dealing with auto-observation (Pitrat, 1993).

## Acknowledgements

We would like to thank Christian Jacquemin, François Pachet, Timothy Pullman and the referees for their helpful comments on earlier drafts of this paper.

## Bibliographical References